\documentclass[10pt,twocolumn,letterpaper]{article}

\usepackage[pagenumbers]{wacv} 

\usepackage{graphicx}
\usepackage{amsmath}
\usepackage{amssymb}
\usepackage{booktabs}
\usepackage[numbers,sort&compress]{natbib}
\usepackage{threeparttable}

\usepackage{algorithm}
\usepackage{algpseudocode}

\usepackage[pagebackref,breaklinks,colorlinks]{hyperref}

\usepackage[capitalize]{cleveref}
\crefname{section}{Sec.}{Secs.}
\Crefname{section}{Section}{Sections}
\Crefname{table}{Table}{Tables}
\crefname{table}{Tab.}{Tabs.}

\def\wacvPaperID{1648} 
\def\confName{WACV}
\def\confYear{2025}

\begin{document}

\title{Enhancing Ecological Monitoring with Multi-Objective Optimization: A Novel Dataset and Methodology for Segmentation Algorithms}

\author{
    \begin{tabular}[t]{c}
        Sophia J. Abraham, Jin Huang, Jonathan D. Hauenstein, Walter Scheirer\\
        University of Notre Dame\\
        {\tt\small \{sabraha2, jhuang24, hauenstein@nd.edu, walter.scheirer\}@nd.edu}
    \end{tabular}
    \and
    \begin{tabular}[t]{c}
        Brandon RichardWebster\\
        Kitware Inc.\\
        {\tt\small brandon.richardwebster@kitware.com}
    \end{tabular}
    \and
    \begin{tabular}[t]{c}
        Michael Milford\\
        Queensland University of Technology\\
        {\tt\small michael.milford@qut.edu.au}
    \end{tabular}
}

\maketitle

\begin{abstract}
We introduce a unique semantic segmentation dataset of 6,096 high-resolution aerial images capturing indigenous and invasive grass species in Bega Valley, New South Wales, Australia, designed to address the underrepresented domain of ecological data in the computer vision community. This dataset presents a challenging task due to the overlap and distribution of grass species, which is critical for advancing models in ecological and agronomical applications. Our study features a homotopy-based multi-objective fine-tuning approach that balances segmentation accuracy and contextual consistency, applicable to various models. By integrating DiceCELoss for pixel-wise classification and a smoothness loss for spatial coherence, this method evolves during training to enhance robustness against noisy data. Performance baselines are established through a case study on the Segment Anything Model (SAM), demonstrating its effectiveness. Our annotation methodology, emphasizing pen size, zoom control, and memory management, ensures high-quality dataset creation. The dataset and code are publicly available [\textbf{link to repo here\footnote{Currently excluded to honor double blind process}}], aiming to drive research in computer vision, machine learning, and ecological studies, advancing environmental monitoring and sustainable development.

\end{abstract}
    
\section{Introduction}
\label{sec:intro}

The global challenge of ensuring food security for a burgeoning population is exacerbated by ecological threats like invasive grass species. These species compromise natural vegetation, pose risks to livestock, and increase wildfire occurrences. The economic burden of managing these invasions, once established, is substantial, costing up to 17 times more than preventative measures \citep{MackRichardN.2000BICE}. A significant contributor to this dilemma is \textit{African lovegrass} (\textit{ALG}), an invasive species that has spread rapidly in both Australia and the U.S. Its robust nature and rapid proliferation demand swift action for control and eradication.

\begin{figure}
\centering
\includegraphics[width=0.5\textwidth,keepaspectratio]{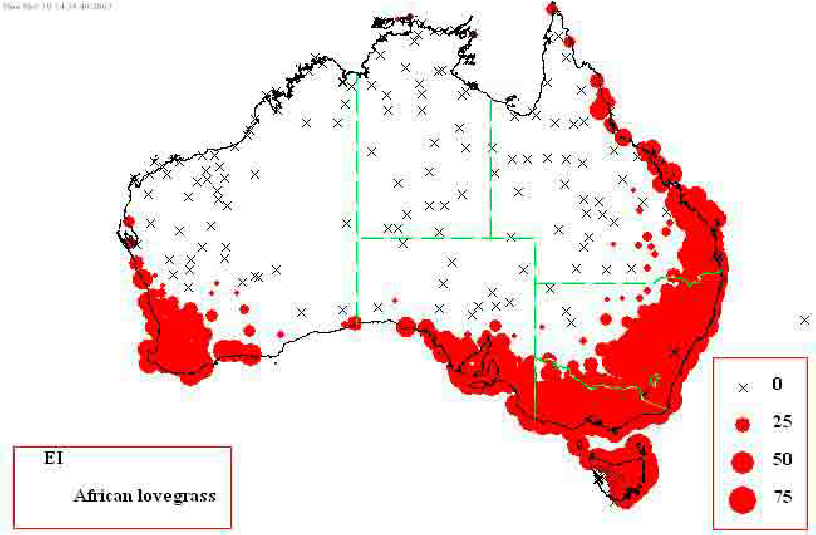}
\caption{Estimated distribution of African lovegrass (ALG) infestation in Australia, based on data from Queensland Primary Industries and Fisheries (2009).}
\label{fig
/algmap}
\end{figure}

Introduced in the 1930s to the Bega Valley, New South Wales, Australia, \textit{ALG} has become a formidable adversary to agricultural and natural landscapes \citep{MackRichardN.2000BICE}. Its hardiness and adaptability to adverse conditions, such as drought and low soil fertility, facilitate its spread across diverse environments. The ecological impact is profound, with \textit{ALG} dominance leading to reduced soil fertility, displacement of native grasses, and a significant decrease in pasture productivity. This increases the risk of wildfires and jeopardizes community safety \citep{Firn2009AfricanLI}. An overview of ALG infestation across Australia can be seen in Fig.~\ref{fig
/algmap}. Current management strategies emphasize early detection and herbicide application at the onset of invasion as critical measures to combat its spread \citep{doi:10.1111/1365-2664.12928}.

In response to the urgent need for effective \textit{ALG} management tools, this paper introduces a novel semantic segmentation dataset consisting of 6,096 high-resolution aerial images. This dataset, focused on capturing indigenous and invasive grass species in the Bega Valley, aims to empower researchers and practitioners with the resources necessary to develop and refine algorithms capable of identifying \textit{ALG} invasions promptly.

This study focuses on the binary task of distinguishing grass from non-grass areas, which is a foundational step towards more fine-grained identification of different grass species, including invasive ones like \textit{ALG}. The binary segmentation task, while already challenging due to the complex and heterogeneous nature of the ecological data, lays the groundwork for future advancements in detailed classification tasks.

This dataset distinguishes itself by offering several unique properties that address common challenges in computer vision and ecological studies:

\textbf{Underrepresented Domain}: Unlike typical datasets focusing on urban environments, vehicles, or human faces, this dataset centers on grass species, a significantly underrepresented area in computer vision.

\textbf{Class Overlap and Distribution}: It captures the complex overlap and distribution of indigenous and invasive grass species, providing a challenging environment for segmentation models.

\textbf{Ecological Relevance}: The dataset includes images taken at various altitudes and conditions, reflecting real-world ecological monitoring scenarios.

This paper also presents an innovative homotopy-based multi-objective fine-tuning approach, demonstrated through a case study on the Segment Anything Model (SAM). Traditional single-objective optimization methods often fall short in addressing the dual demands of precise segmentation and contextual coherence, especially in noisy and heterogeneous ecological data. Our approach dynamically balances segmentation accuracy and contextual consistency by integrating DiceCELoss for precise pixel-wise classification and a smoothness loss to ensure spatial coherence. The homotopy parameter evolves during training, enabling a smooth transition from prioritizing segmentation accuracy to emphasizing contextual consistency. This dual-objective strategy enhances the robustness and reliability of segmentation results. While this study focuses on SAM, our approach is general enough to be applicable to other segmentation models as well.

Through rigorous evaluation, we establish performance baselines for our fine-tuned SAM and compare it with other leading semantic segmentation models. These baselines highlight the dataset's potential to advance machine learning techniques in ecological monitoring. Our annotation methodology further enriches the dataset, providing valuable insights for researchers involved in dataset creation.

By providing a robust dataset and an advanced fine-tuning approach, this work aims to drive forward the fields of environmental monitoring and sustainable development, fostering innovations that can address complex ecological challenges and safeguard agricultural productivity and ecosystem health. This dataset not only serves as a practical tool for addressing the specific challenge of \textit{ALG} but also represents a broader contribution to the computer vision community by presenting a unique and challenging dataset for model development and evaluation.

\section{Related Works}
\label{sec:related_work}

Despite existing advancements in identifying invasive plant species, the application of these techniques to aerial imagery, particularly for species such as African lovegrass (ALG), remains largely unexplored \cite{4d0d0a2d4f784d0495d69625c02293b9,10.1371/journal.pone.0031734,bradley2014,goeau2016,su12093544}. The remote sensing community's recent ventures into leveraging remote sensing methodologies for invasive plant detection mark a significant step forward, yet these efforts rarely extend to grass-specific identification \cite{ismail2016identification,PMID:30045513,bolch2020remote}. Notably, Albani et al.'s work on estimating weed coverage in open fields through UAVs underscores the potential of aerial technologies, though it stops short of achieving species-level weed identification \cite{albani2017}.

While considerable research has been directed towards weed identification within agricultural contexts \cite{PATRICIO201869,10.1093/gigascience/giy153,7943593,TIAN20201,chandra2020computer}, these studies predominantly focus on irrigated soils, overlooking the unique challenges presented by open prairie environments. Compounding the issue is the inherent difficulty in distinguishing among a vast array of plant species, exacerbated by similarities in color and shape, as highlighted by Wäldchen et al. \cite{https://doi.org/10.1111/2041-210X.13075}. Their findings advocate for an interdisciplinary approach, merging the expertise of biologists and computer scientists to propel forward the field of plant identification.

This interdisciplinary proposition raises the intriguing possibility of developing a robust algorithm for ALG detection that integrates insights from ecology, computer vision, and visual psychophysics. Given the nascent state of research into the semantic segmentation of prairie grass, a critical preliminary step involves evaluating the efficacy of existing semantic segmentation methodologies \cite{long2015fully,chen2017rethinking,badrinarayanan2017segnet,10.1007/978-3-319-24574-4_28,farabet2012learning,socher2011parsing,noh2015learning,zheng2015conditional,chen2017deeplab} in this novel context.

\section{Multi-Objective Methodology}
\label{sec:methodology}

In this section, we detail the homotopy-based multi-objective fine-tuning approach applied to the Segment Anything Model (SAM). This method addresses the dual objectives of segmentation accuracy and contextual consistency, which are crucial for handling noisy and heterogeneous ecological data.

Our approach aims to optimize two primary objectives:
\begin{enumerate}
    \item \textbf{Segmentation Accuracy}: Ensured by DiceCELoss \cite{sudre2017generalised}, which provides precise pixel-wise classification.
    \item \textbf{Contextual Consistency}: Achieved through a smoothness loss that promotes spatial coherence across segmentations.
\end{enumerate}

DiceCELoss combines Dice Loss and Cross-Entropy Loss
in order to capture both the overlap between predicted and ground truth masks as well as the pixel-wise classification accuracy.

The Dice Loss is defined as:
\begin{equation}
L_{\text{Dice}} = 1 - \frac{2 \sum_{i=1}^{N} p_i g_i + \epsilon}{\sum_{i=1}^{N} p_i + \sum_{i=1}^{N} g_i + \epsilon}
\end{equation}
where \( p_i \) and \( g_i \) are the predicted and ground truth labels for pixel \( i \), respectively, and \( \epsilon \) is a small positive constant to avoid division by zero.

The Cross-Entropy Loss is defined as:
\begin{equation}
L_{\text{CE}} = - \frac{1}{N} \sum_{i=1}^{N} \left[ g_i \log(p_i) + (1 - g_i) \log(1 - p_i) \right]
\end{equation}

The combined DiceCELoss is given by:
\begin{equation}
L_{\text{DiceCE}} = \beta L_{\text{Dice}} + (1 - \beta) L_{\text{CE}}
\end{equation}
where \( \beta \) is a weighting factor between 0 and 1.

To ensure contextual consistency, we incorporate a smoothness loss \cite{rother2004grabcut} that penalizes abrupt changes in the segmentation map. The smoothness loss is defined as:
\begin{equation}
\begin{aligned}
L_{\text{smooth}} = \lambda_{\text{smooth}} \Bigg( & \sum_{i=1}^{N-1} \sum_{j=1}^{M} \left| p_{i,j} - p_{i+1,j} \right| \\
+ & \sum_{i=1}^{N} \sum_{j=1}^{M-1} \left| p_{i,j} - p_{i,j+1} \right| \Bigg)
\end{aligned}
\end{equation}
where \( p_{i,j} \) represents the predicted label at pixel \((i,j)\), and \( \lambda_{\text{smooth}} \) is a regularization parameter.

\subsection{Homotopy-Based Multi-Objective Optimization}

Homotopy methods provide a systematic way to transition between different objective functions during optimization. In our approach (Algorithm \ref{alg:homotopy_optimization}), we dynamically balance the two objectives by introducing a homotopy parameter \( t \) that evolves from 0 to 1 over the course of training. This allows the optimization process to start by focusing on segmentation accuracy and gradually shift towards emphasizing contextual consistency as training progresses.

We define the combined loss function as:
\begin{equation}
L_{\text{combined}} = (1 - t) L_{\text{DiceCE}} + t L_{\text{smooth}}
\end{equation}
where \( t \) smoothly transitions the objective from prioritizing segmentation accuracy (when \( t = 0 \)) to prioritizing smoothness (when \( t = 1 \)). This gradual shift helps in preventing the model from overfitting to one objective too early and ensures a balanced optimization process \citep{allgower2012numerical}.

The overall training objective is to minimize the combined loss function:
\begin{equation}
\mathcal{L} = \min \left( (1 - t) L_{\text{DiceCE}} + t L_{\text{smooth}} \right)
\end{equation}

\subsection{Training Procedure}
The training procedure is as follows:
\begin{enumerate}
    \item \textbf{Initialization}: Initialize SAM with pre-trained weights.
    \item \textbf{Data Augmentation}: Apply data augmentation techniques to enhance the diversity of the training dataset.
    \item \textbf{Optimization}: For each epoch, compute the combined loss and update the model parameters using gradient descent. Adjust the homotopy parameter \( t \) to gradually shift the focus from segmentation accuracy to contextual consistency.
    \item \textbf{Evaluation}: Evaluate the model on a validation set to monitor performance and adjust hyperparameters as needed.
\end{enumerate}

\begin{algorithm}[t]
\caption{Homotopy-Based Fine-Tuning for SAM}
\label{alg:homotopy_optimization}
\begin{algorithmic}[1]
\Require $\mathcal{D}$ (dataset), $\mathcal{M}$ (SAM model), $\alpha_{\text{start}}$ (start learning rate), $\alpha_{\text{end}}$ (end learning rate), $T$ (total training steps)
\Ensure Optimized SAM model $\mathcal{M}^*$
\State Initialize model $\mathcal{M}$ with pre-trained weights
\State Initialize learning rate $\alpha \gets \alpha_{\text{start}}$
\State Initialize homotopy parameter $t \gets 0$
\For{$step \gets 1$ to $T$}
    \State Sample a batch $\{(\mathbf{x}_i, \mathbf{y}_i)\}_{i=1}^N \sim \mathcal{D}$
    \State Compute predictions $\hat{y}_i \gets \mathcal{M}(\mathbf{x}_i)$
    \State Compute DiceCELoss $L_{\text{DiceCE}}$
    \State Compute Smoothness Loss $L_{\text{smooth}}$
    \State Compute combined loss $L_{\text{combined}}$
    \State Update model $\mathcal{M}$ using gradients of $L_{\text{combined}}$
    \State Update learning rate $\alpha$ and homotopy parameter $t$
    \State $\alpha \gets \text{LinearSchedule}(\alpha_{\text{start}}, \alpha_{\text{end}}, T, step)$
    \State $t \gets \frac{step}{T}$
\EndFor
\State $\mathcal{M}^* \gets \mathcal{M}$
\State \Return $\mathcal{M}^*$
\end{algorithmic}
\end{algorithm}

\section{ALGSeg}
\label{algseg}

\begin{figure}[t]
 \centering
    \includegraphics[width=0.45\textwidth,keepaspectratio]{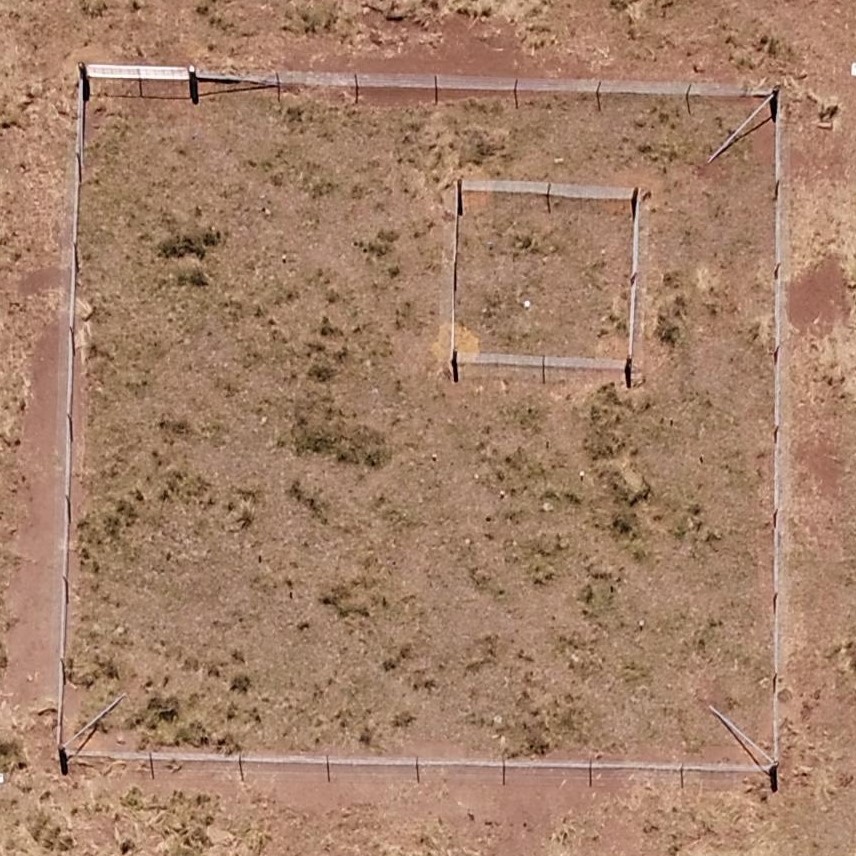}
 \caption{A 25x25m$^2$ plot established for ecological monitoring, including the study of native grasses and African lovegrass interactions.}
 \label{fig:ch6/plot-25x25}
\end{figure}

In an interdisciplinary collaboration, our team comprising drone operators, ecologists, and roboticists undertook a field study in the Bega Valley, New South Wales, Australia, during the late Australian spring in November. Our objective was to establish a comprehensive dataset for monitoring ecological changes, with a particular focus on the dynamics between native grasses and the invasive African lovegrass (ALG). We selected and set up more than twenty 25x25m plots, aimed at long-term ecological monitoring over the next five years. These plots are designated for tracking changes in insect populations, soil nutrient content, and the composition of plant and grass species, including the interaction between native grasses and ALG. An example of one of these plots is illustrated in Fig.~\ref{fig:ch6/plot-25x25}.

Utilizing aerial photography, we began the process of automating the classification of grass species from drones. This initiative is aimed at assisting farmers and landcare workers in monitoring the spread of ALG through a shared platform that integrates the classification data with GPS coordinates. Our efforts concentrated on Merimbula, a pivotal location within the Bega Valley region. 

\subsection{Data Collection}

Aerial data collection was conducted over private lands, with the gracious permission and enthusiasm of the landowners situated in the vicinity of Merimbula. For the purpose of maintaining anonymity while allowing for clear reference, the properties were designated as lots L1, L2, and L3, detailed in the supplemental. This coding system ensures the privacy of landowner identities while facilitating structured data analysis.

Over the initial two days of the project, aerial imaging captured distinct areas, denoted as 'a' and 'b' for each lot, to provide comprehensive coverage. Weather conditions varied significantly during the data collection period, ranging from a combination of cloudy and sunny skies on the first day to overcast conditions on the second, and predominantly sunny weather on the third day. These variations in lighting conditions are meticulously documented in the supplemental, highlighting the adaptability of our data collection process to environmental changes. For detailed data collection specifics, please refer to the supplemental materials.

\textit{Note}: The careful consideration of weather conditions and the collaboration with local landowners not only exemplifies the logistical planning required for remote sensing projects but also underscores the importance of community engagement in scientific research.

\subsubsection{Equipment Used for Recording}
Aerial imagery was captured using a DJI Inspire 2 equipped with a Zenmuse X5S camera, which boasts a resolution of 20 megapixels (5280x3956 pixels). The camera was set to capture still shots at two-second intervals, employing JPEG lossy compression at the standard compression ratio. This setup ensured an approximate 75\% overlap both forward and to the sides of each image, facilitating the subsequent processing of these images into orthomosaics.

\subsubsection{Collection of Auxiliary Data}
Beyond the primary dataset, we compiled a comprehensive set of auxiliary data to support a broad spectrum of computer vision and robotics research. This dataset encompasses extensive real-time flight metrics, including but not limited to, operational parameters of the drone (e.g., flight times, durations, distances, and varied flight attitudes). It is important to note that to uphold privacy standards, these flight details are currently undergoing an anonymization process to prevent any potential compromise of data integrity before their public release.

Additionally, while the drone's onboard systems do not capture meteorological data, we have supplemented our dataset with detailed weather information corresponding to each flight, courtesy of the Australian Government's Bureau of Meteorology. This supplementary weather dataset offers a rich array of atmospheric conditions surrounding each flight, providing variables such as temperature, dew point, humidity, wind patterns, and precipitation. The integration of these weather parameters, which are distinct from the settings used by drone operators for white balance adjustments, enriches the dataset with environmental context crucial for certain computer vision and robotics applications.

\textit{Note:} The weather information serves to complement the visual data, offering insights into environmental conditions that may influence the analysis and application of the collected imagery in various computational models.

\subsection{Data Annotation}
In creating a dataset with high-resolution images that capture a wide variety of content, the annotation process emerges as a formidable challenge, particularly when it comes to the nuanced details of various grass types and their phenotypic stages—observable characteristics at different growth phases. Given the practical constraints at the time of data collection, enlisting experts for detailed annotation was not feasible. This led us to devise a simplified annotation strategy aimed at distinguishing between grass and non-grass elements, a method that directly tackles the inherent complexities of natural prairie imagery where a single pixel might fall into multiple categories, such as grass, other vegetation, objects, or soil. This streamlined approach, while reducing the scope of the task, nonetheless required considerable effort, with annotation times ranging from 45 minutes to 2 hours per image. For this study, we curated a set of 50 images for annotation to ensure a diverse representation across all sampled locations.

Addressing the broader challenge, the reliance on non-expert annotators due to the unavailability of experts opens a dialogue on the feasibility and limitations of employing laypersons for certain annotation tasks. This scenario, reflective of real-world constraints, offers valuable insights into the adaptability of machine learning applications in the face of resource limitations. By navigating these constraints, our approach contributes to the discourse on developing efficient, scalable solutions for dataset creation in the field of Earth Observation. Such explorations into the capabilities of non-expert annotators underscore the importance of innovative methodologies in enhancing data annotation practices, paving the way for broader applications and understandings within the domain.

\textit{Note:} For further details on the specific flight parameters, image capture elevations, and examples of artifacts resulting from the annotation process, please refer to the supplemental materials.

\section{Baselines and Segmentation Models}
\label{sec:baselines}

In this study, we explore pixel-wise semantic segmentation for prairie grass by establishing benchmarks using various deep learning-based segmentation techniques. Given the absence of models specifically designed for grass segmentation in prairie settings, we selected models based on their architectural suitability and performance potential for our dataset. This section outlines the chosen models and describes the training regime implemented for their evaluation.

\subsection{Model Selection and Architecture}

We selected a diverse set of models, each with unique architectural features promising for semantic segmentation tasks:

\textbf{DeepLabV3:} The DeepLabV3 models, including versions with 50 and 101 layers, leverage atrous convolution and atrous spatial pyramid pooling to capture multi-scale features. These models balance the trade-off between the number of parameters and feature resolution control. DeepLabV3-101, with a deeper ResNet architecture compared to its 50-layer counterpart, aims to extract more complex features, albeit with increased computational requirements \cite{chen2017rethinking}.

\textbf{FCN ResNet:} Fully Convolutional Networks (FCNs) extended onto ResNet architectures (50 and 101 layers) utilize ImageNet-pretrained weights for semantic segmentation. These models resize feature maps to match input dimensions, with deeper versions capturing more complex feature hierarchies for potentially improved performance \cite{long2015fully}.

\textbf{SegNet:} SegNet employs an autoencoder architecture with efficient non-linear sparse upsampling in its decoder. Originally optimized for indoor and traffic segmentation tasks, SegNet's architecture emphasizes computational efficiency \cite{badrinarayanan2017segnet}.

\textbf{U-Net:} Known for its effectiveness across various semantic segmentation challenges, U-Net employs a downsample-then-upsample approach with skip connections. This architecture facilitates the fusion of multi-level feature information, making it robust for diverse segmentation tasks \cite{10.1007/978-3-319-24574-4_28}.

\textbf{Segment Anything Model (SAM):} In addition to these established models, we include the Segment Anything Model (SAM) in two configurations: single-objective and our proposed multi-objective fine-tuning approach. The single-objective SAM focuses on segmentation accuracy, while the multi-objective SAM balances segmentation accuracy with contextual consistency using the homotopy-based approach \cite{kirillov2023segment}.

\subsection{Training Regime}

The training and evaluation of these models were conducted on the African Lovegrass (ALG) dataset, segmented into 224x224px patches to ensure uniform input sizes. The dataset was split into a 90/10 ratio, resulting in 19,440 training patches and 2,160 evaluation patches. This split aimed to avoid overfitting and provide a robust assessment of each model's capabilities.

\textbf{Initialization and Training:} All models were initialized with pretrained weights where available, except for U-Net and SegNet. Each model underwent a standardized training process over 250 epochs using the default configurations for loss functions, optimizers, and batch sizes. The training was performed within the flexible PyTorch framework, accommodating the specific input dimensions.

For SAM, the fine-tuning process was tailored to balance segmentation accuracy and contextual consistency through a homotopy-based approach. The SAM model was initialized with pre-trained weights from sam-vit-base. The training process involved 100 epochs, where the combined loss function integrated DiceCELoss for pixel-wise classification and a smoothness loss to ensure spatial coherence. The homotopy parameter \( t \) evolved linearly from 0 to 1 over the training epochs, facilitating a smooth transition from prioritizing segmentation accuracy to emphasizing contextual consistency. The optimizer used was Adam with a learning rate of \(1 \times 10^{-5}\).

\textbf{Data Augmentation:} To enhance model robustness, we applied data augmentation techniques such as rotations, flips, and color jittering. This helped in creating a more diverse training set and mitigating overfitting.

\textbf{Evaluation Metrics:} Model performance was evaluated using standard metrics such as Intersection over Union (IoU) and F1-score. These metrics provided a comprehensive understanding of each model's segmentation accuracy and generalization capability.

By establishing these baselines and introducing SAM with both single and multi-objective fine-tuning, we aim to provide a comprehensive evaluation framework for future research in ecological and agronomical segmentation tasks.

\section{Experiments}
\label{sec:experiments}

\begin{figure}[ht!]
 \centering
 \includegraphics[width=0.5\textwidth]{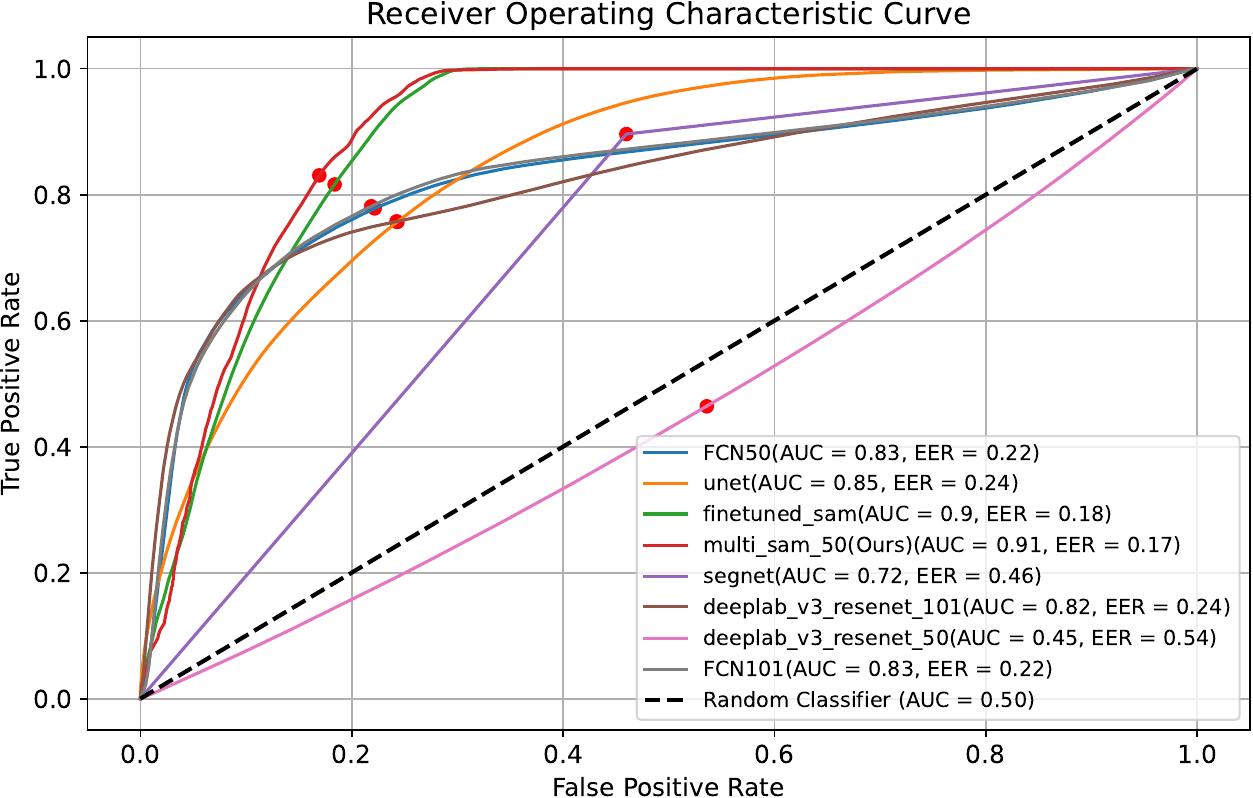}
 \caption{Each ROC curve represents a model's ability to distinguish between grass and non-grass areas, with the Area Under the Curve (AUC) and Equal Error Rate (EER) values indicated in the legend. The EER points are marked with red circles on each curve, showing the threshold where the false positive rate equals the false negative rate. The dashed diagonal line represents the performance of a random classifier (AUC = 0.50). The higher AUC and lower EER values indicate better performance.}
 \label{fig:roc}
\end{figure}

Following the completion of the training phase for each model, we embarked on evaluating their performance using a set of predetermined metrics. The evaluation process began with the generation of Receiver Operating Characteristic (ROC) curves for each model, utilizing the predicted masks from both training and validation datasets to identify the optimal threshold. This threshold was determined by approximating the Equal Error Rate (EER), providing a balanced starting point for further adjustments based on specific user requirements, such as a preference for false positives by landowners (Fig.~\ref{fig:roc}).

The models' output, initially on different scales due to varying final scalar functions (e.g., Sigmoidal), was normalized to the $[0,1]$ range for accurate threshold application. The scaling parameters used for normalizing the training/validation data were also applied to the test data's predicted masks to maintain consistency.

To comprehensively assess the models' baseline performance, we employed the following key metrics:

\textbf{Accuracy} measures the proportion of correctly predicted pixels over the total pixel count, considering true positives, true negatives, false positives, and false negatives.
    
The \textbf{Jaccard Index (IoU)} calculates the overlap area between the predicted and target segmentations relative to their union area, providing a measure of the segmentation's accuracy.
    
The \textbf{DICE score (F1 score)} refines the Jaccard Index by considering twice the overlap area divided by the sum of pixels in both predicted and target images, offering a balanced measure of precision and recall.
    
\textbf{ROC-AUC} (Receiver Operating Characteristic - Area Under Curve) quantifies the overall ability of the model to discriminate between classes, providing a single measure of model performance across all classification thresholds.
    
\textbf{EER} (Equal Error Rate) is the point at which the false positive rate equals the false negative rate, providing a balanced measure of the model's accuracy at a specific threshold.

These metrics provide a comprehensive evaluation of model performance, highlighting strengths and weaknesses in segmentation accuracy, contextual coherence, and overall discriminative ability.

\section{Results}
\begin{table*}[h]
\centering
\caption{Performance Metrics for Different Segmentation Models}
\label{tab:performance_metrics}
\begin{tabular}{lrrrrrrr}
\toprule
\textbf{Model Name} & \textbf{Accuracy $\uparrow$} & \textbf{Jaccard $\uparrow$} & \textbf{Dice $\uparrow$} & \textbf{ROC AUC $\uparrow$} & \textbf{EER $\downarrow$} & \textbf{EER Threshold} \\
\midrule
FCN50 & 0.67 & 0.63 & 0.77 & 0.83 & 0.22 & 0.54  \\
Unet & 0.75 & 0.52 & 0.68 & 0.85 & 0.24 & 0.44  \\
Finetuned SAM & 0.89 & 0.67 & 0.80 & 0.90 & 0.18 & 0.70  \\
\textbf{Multi SAM 50 (Ours)} & \textbf{0.90} & 0.67 & 0.80 & \textbf{0.91} & \textbf{0.17} & 0.74  \\
SegNet & 0.77 & 0.61 & 0.76 & 0.72 & 0.46 & 1.00 \\
DeepLabV3 ResNet 101 & 0.75 & 0.54 & 0.70 & 0.82 & 0.24 & 0.50 \\
DeepLabV3 ResNet 50 & 0.65 & 0.65 & 0.79 & 0.45 & 0.54 & 0.51  \\
FCN101 & 0.67 & 0.63 & 0.77 & 0.83 & 0.22 & 0.55  \\
\bottomrule
\end{tabular}
\end{table*}
In this section, we analyze the performance of various segmentation models on our novel ecological dataset, focusing on the task of binary grass segmentation. The dataset consists of high-resolution aerial images captured at different altitudes (10m, 35m, and 120m) over the Bega Valley in Australia. 

Multi SAM 50 (Ours) model achieved the highest accuracy (0.90) and ROC AUC (0.91), along with the lowest EER (0.17). The Dice Score (0.80) and Jaccard Index (0.67) were also among the best. The strong performance of Multi SAM 50 underscores the effectiveness of our homotopy-based multi-objective fine-tuning approach. By optimizing for both segmentation accuracy and contextual consistency, this model balances precision and robustness. Notably, it achieved these results with only 50 epochs of training, compared to the 100 epochs required for the single-objective Finetuned SAM, demonstrating the efficiency of our method.

Finetuned SAM also performed well, with an accuracy of 0.89, ROC AUC of 0.90, and EER of 0.18. The Dice Score (0.80) and Jaccard Index (0.67) were comparable to those of Multi SAM 50. The high performance of Finetuned SAM indicates the robustness of the SAM architecture for segmentation tasks. However, the additional smoothness objective in Multi SAM 50 provides a clear advantage in balancing segmentation accuracy with contextual consistency.

SegNet achieved an accuracy of 0.77, Jaccard Index of 0.61, and Dice Score of 0.76. However, its ROC AUC (0.72) and EER (0.46) were less favorable. The relatively high EER indicates that SegNet may produce more false positives, which is less desirable for applications requiring precise segmentation. Its performance highlights the importance of choosing models that balance false positives and negatives effectively.

DeepLabV3 ResNet 101 had an accuracy of 0.75, Jaccard Index of 0.54, Dice Score of 0.70, ROC AUC of 0.82, and EER of 0.24. The ResNet 50 variant showed slightly lower performance metrics. These models demonstrated consistent performance but were outperformed by both SAM-based models and Multi SAM 50. The results suggest that while DeepLabV3 models are robust, they may require additional tuning or different pretraining strategies to match the performance of SAM-based approaches.

 FCN50 and FCN101 achieved similar metrics, with accuracies of 0.67, Jaccard Indices of 0.63, Dice Scores of 0.77, ROC AUCs of 0.83, and EERs of 0.22. These models showed consistent performance but were outperformed by newer architectures like SAM and Multi SAM 50.

 The ROC curve and performance metrics table provide a multi-faceted view of each model's performance. Some key observations include:

\textbf{Invariance to Tuning}: Different models exhibit varying levels of sensitivity to tuning parameters. For instance, the Multi SAM 50 model, fine-tuned with only 50 epochs, achieved superior performance compared to the Finetuned SAM, which required 100 epochs. This indicates that our multi-objective approach may provide a more efficient optimization path.

\textbf{Altitude Variations}: The dataset's multi-altitude nature (10m, 35m, 120m) adds complexity to the segmentation task. Models must generalize across different resolutions and perspectives, which can affect performance. Further analysis could explore how models trained at specific altitudes perform compared to those trained on mixed-altitude data.

\textbf{Trade-offs in Performance}: The high precision end of the ROC curves reveals interesting trade-offs. For instance, models with higher AUC and lower EER values offer better recall at low error rates, making them suitable for applications where minimizing false positives is crucial. Conversely, models like SegNet, which perform well in terms of Dice and Jaccard scores, may be better for tasks requiring high segmentation precision within correctly classified areas.

\textbf{Generalization to New Environments}: The varying performance across models suggests differences in their ability to generalize to unseen environments. Models with lower EER and higher AUC are likely more robust to out-of-distribution data, an essential factor for practical deployment in diverse ecological settings.


\begin{figure*}[ht!]
    \centering
    \begin{subfigure}[b]{0.19\textwidth}
        \centering
        \includegraphics[width=\linewidth]{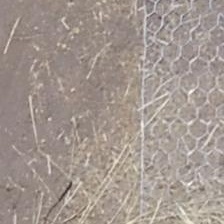}
        \caption{Aerial Image}
        \label{fig:aerial}
    \end{subfigure}
    \hfill
    \begin{subfigure}[b]{0.19\textwidth}
        \centering
        \includegraphics[width=\linewidth]{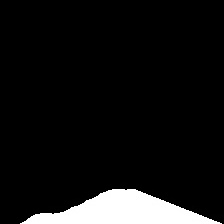}
        \caption{Ground Truth Mask}
        \label{fig:mask}
    \end{subfigure}
    \hfill
    \begin{subfigure}[b]{0.19\textwidth}
        \centering
        \includegraphics[width=\linewidth]{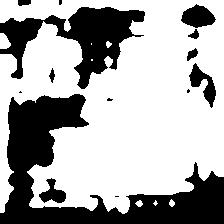}
        \caption{DeepLabV3 ResNet 50}
        \label{fig:deeplab50}
    \end{subfigure}
    \hfill
    \begin{subfigure}[b]{0.19\textwidth}
        \centering
        \includegraphics[width=\linewidth]{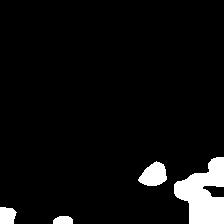}
        \caption{DeepLabV3 ResNet 101}
        \label{fig:deeplab101}
    \end{subfigure}
    \vspace{0.3cm}
    \begin{subfigure}[b]{0.19\textwidth}
        \centering
        \includegraphics[width=\linewidth]{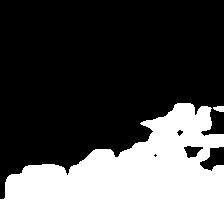}
        \caption{FCN50}
        \label{fig:fcn50}
    \end{subfigure}
    \hfill
    \begin{subfigure}[b]{0.19\textwidth}
        \centering
        \includegraphics[width=\linewidth]{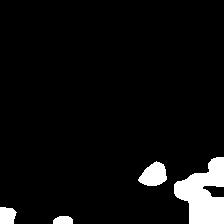}
        \caption{FCN101}
        \label{fig:fcn101}
    \end{subfigure}
    \hfill
    \begin{subfigure}[b]{0.19\textwidth}
        \centering
        \includegraphics[width=\linewidth]{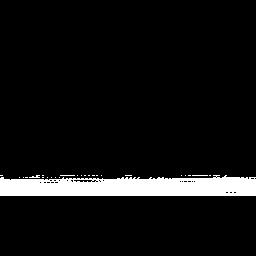}
        \caption{Finetuned SAM}
        \label{fig:sam}
    \end{subfigure}
    \hfill
    \begin{subfigure}[b]{0.19\textwidth}
        \centering
        \includegraphics[width=\linewidth]{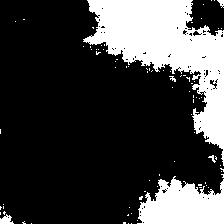}
        \caption{SegNet}
        \label{fig:segnet}
    \end{subfigure}
    \vspace{0.3cm}
    \begin{subfigure}[b]{0.19\textwidth}
        \centering
        \includegraphics[width=\linewidth]{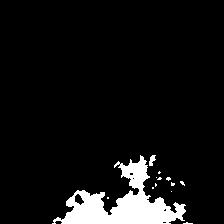}
        \caption{U-Net}
        \label{fig:unet}
    \end{subfigure}
    \hfill
    \begin{subfigure}[b]{0.19\textwidth}
        \centering
        \includegraphics[width=\linewidth]{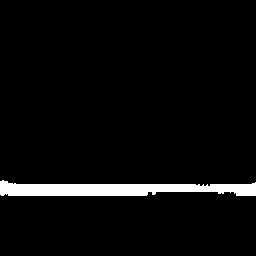}
        \caption{Multi-Objective SAM}
        \label{fig:multiobj}
    \end{subfigure}
    \caption{Comparison of segmentation results from all models on the same aerial image. The comparison highlights the varying levels of detail, precision, and generalization ability of each model in distinguishing grass from non-grass areas. Note that some models like Multi-Objective SAM tend to misclassify large portions of the image as grass, while others like DeepLabV3 ResNet 101 perform better at capturing fine details.}
    \label{fig:segmentation_comparison}
\end{figure*}

\section{Limitations}
In our error analysis (samples visualized in Figure \ref{fig:segmentation_comparison}), we identified several common types of errors observed in the segmentation results. Models often struggled with:

\textbf{Boundary Errors}: Misclassifications at the edges of grass patches, where the transition between grass and non-grass is abrupt.

\textbf{Small Patch Detection}: Difficulty in accurately segmenting small grass patches due to limited resolution and feature representation.

\textbf{False Positives and Negatives}: Instances where non-grass areas were misclassified as grass and vice versa, often due to similarities in texture or color.

These errors can be attributed to several factors, including the complexity of the ecological data, variations in lighting conditions, and the inherent difficulty in distinguishing between visually similar textures. Future work will focus on addressing these challenges by incorporating more sophisticated models and additional training data to improve segmentation accuracy and robustness.

\section{Conclusion}
\label{conclusion}
This study has demonstrated the feasibility and challenges of the binary task of distinguishing grass from non-grass in aerial images, which serves as a foundational step towards more fine-grained identification of different grass species, including invasive ones like African lovegrass (ALG). Our research highlights the complexity of accurately segmenting ecological data due to variations in texture, lighting, and the presence of other vegetation.

The identification of common segmentation errors, such as boundary inaccuracies and false positives/negatives, provides critical insights for improving future models. Enhancing these models through better algorithms and more comprehensive training data will increase their robustness and reliability. The multi-objective approach, which balances segmentation accuracy with contextual consistency, shows promise in mitigating these errors. The strong performance of the Multi-Objective SAM model validates the effectiveness of our homotopy-based multi-objective fine-tuning approach. By optimizing both segmentation accuracy and contextual consistency, the Multi-Objective SAM model not only achieves high precision but also demonstrates efficiency, requiring only 50 epochs compared to the 100 epochs for the single-objective Finetuned SAM.

The implications of this work extend to practical applications in ecological monitoring and management. Accurate segmentation of ALG can significantly enhance the efficiency of monitoring and managing invasive species, potentially reducing their economic and ecological impact. The high accuracy and low Equal Error Rate (EER) of the Multi-Objective SAM model indicate that advanced segmentation techniques can provide reliable data for early detection and proactive management of invasive species.

While this study is focused on binary segmentation, it sets the stage for future work aimed at fine-grained classification tasks. The insights and methodologies developed here will be instrumental in advancing the capabilities of segmentation models, ultimately contributing to more precise and effective ecological management strategies.

{\small
\bibliographystyle{ieee_fullname}
\bibliography{PaperForReview}
}

\end{document}